\documentclass[fleqn,10pt]{wlscirep}
\usepackage[utf8]{inputenc}
\usepackage[T1]{fontenc}

\usepackage{adjustbox,lipsum} 
\newcommand{\bftab}{\fontseries{b}\selectfont}
\usepackage{caption}
\usepackage{subcaption}
\usepackage{xurl} 
\usepackage{multirow} 

\usepackage{pifont} 
\newcommand{\xmark}{\ding{55}} 

\title{RescueNet: A High Resolution UAV Semantic Segmentation Dataset for Natural Disaster Damage Assessment}

\author[1,2,*]{Maryam Rahnemoonfar}
\author[3]{Tashnim Chowdhury}
\author[4]{Robin Murphy}
\affil[1]{ Department of Computer Science and Engineering, Lehigh University, Bethlehem, Pennsylvania, 18015, USA}
 \affil[2]{Department of Civil and Environmental Engineering, Lehigh University, Bethlehem, Pennsylvania, 18015, USA}
\affil[3]{Department of Information Systems, University of Maryland, Baltimore County, Baltimore, Maryland, 21250, USA}
\affil[4]{Department of Computer Science and Engineering, Texas A \& M University, College Station, Texas, 77843, USA}

\affil[*]{corresponding author(s): Maryam Rahnemoonfar (maryam@lehigh.edu)}

\begin{abstract}

Recent advancements in computer vision and deep learning techniques have facilitated notable progress in scene understanding, thereby assisting rescue teams in achieving precise damage assessment. In this paper, we present RescueNet, a meticulously curated high-resolution post-disaster dataset that includes detailed classification and semantic segmentation annotations. This dataset aims to facilitate comprehensive scene understanding in the aftermath of natural disasters. RescueNet comprises post-disaster images collected after Hurricane Michael, obtained using Unmanned Aerial Vehicles (UAVs) from multiple impacted regions. The uniqueness of RescueNet lies in its provision of high-resolution post-disaster imagery, accompanied by comprehensive annotations for each image. Unlike existing datasets that offer annotations limited to specific scene elements such as buildings, RescueNet provides pixel-level annotations for all classes, including buildings, roads, pools, trees, and more. Furthermore, we evaluate the utility of the dataset by implementing state-of-the-art segmentation models on RescueNet, demonstrating its value in enhancing existing methodologies for natural disaster damage assessment. The dataset can be downloaded from \url{https://www.dropbox.com/scl/fo/ntgeyhxe2mzd2wuh7he7x/AHJ-cNzQL-Eu04HS6bvBgcw?rlkey=6vxiaqve9gp6vzvzh3t5mz0vv&e=1&dl=0}

\end{abstract}
\begin{document}

\flushbottom
\maketitle

\thispagestyle{empty}

\section*{Background \& Summary}

\par In recent years, natural disasters have profoundly impacted various regions across the globe, primarily attributed to the escalating effects of climate change and other contributing factors. The impacts of these disasters are getting stronger and more lasting. Reducing economic losses and saving valuable human lives depends heavily on quick response from the rescue teams. Various computer vision techniques can significantly contribute to precise damage assessment by leveraging the visual elements inherent in imagery.
Image classification, object detection, instance segmentation, and semantic segmentation represent distinct components within the field of computer vision.. Image classification is the task of assigning a label or class to an entire image. Object detection is another computer vision technique that works to identify and locate objects within an image or video. Specifically, object detection draws bounding boxes around these detected objects, which help to locate where said objects are in a given scene. Semantic segmentation is one of the most essential tools for scene understanding. Semantic segmentation performs pixel level classification of each object of an image with distinct boundaries. On the other hand, instance segmentation is a unique form of image segmentation that deals with detecting and delineating each distinct instance of an object appearing in an image. Recently, substantial progress has been achieved in the field of scene understanding in urban environments, primarily attributable to the application of deep learning methodologies. Several pioneering datasets like Cityscapes \cite{cordts2016cityscapes}, PASCAL VOC2012 \cite{everingham2010pascal}, PASCAL Context \cite{mottaghi2014role}, and COCO Stuff dataset \cite{caesar2018coco} are available and these large-scale datasets are helping in improvement of segmentation accuracy of urban scenes. Despite these advances in semantic segmentation, scene understanding after natural disaster remains challenging due to the absence of benchmark datasets.

\par Existent natural disaster datasets can be sorted into two classes. One is ground-level images \cite{nguyen2017damage, nia2017building, weber2020detecting}, and other one is satellite and aerial imagery \cite{chen2018benchmark, gupta2019creating, rahnemoonfar2021floodnet}. The ground level images are collected from different sources including Virtual Disaster Viewer \cite{nia2017building}, Open Images Dataset \cite{nia2017building}, Google image search engine \cite{nia2017building, weber2020detecting}, and social networks \cite{nguyen2017damage, nguyen2017automatic}. Weber \textit{et al.} presented a large ground level post-disaster dataset for different incident classification like drought, wildfire, and snowstorm \cite{weber2020eccv}. Images had been introduced to the AIDR system \cite{nguyen2017automatic} by Nguyen \textit{et al.} by collecting images from the social media \cite{nguyen2017damage}. Although these imageries are abundant, they lack geo location tags \cite{zhu2020msnet} and only provide classification labels (one label for entire image).

\par On the other hand, satellite and aerial imagery are collected from different satellites and remote sensing equipment. Notable works in this arena includes a post-tsunami aerial image dataset named ABCD (AIST Building Change Detection) \cite{fujita2017damage}. This dataset helps in detection of flood affected buildings. Doshi \textit{et al.} proposed a dataset \cite{doshi2018satellite} that combines SpaceNet \cite{cosmiqworksnvidia} and DeepGlobe \cite{demir2018deepglobe}. In order to assess the damages caused by hurricanes, Chen \textit{et al.} proposed a dataset \cite{chen2018benchmark} by collecting images from two different sources including crowdsourced annotated data of DigitalGlobe satellite imagery and data collected by FEMA. A large dataset named xBD which consists of both pre- and post-disaster satellite images are proposed by Gupta \textit{et al.} \cite{gupta2019creating} for building damage assessment with four different damage categories. ISBDA (Instance Segmentation in Building Damage Assessment) is presented by Zhu \textit{et al.} \cite{zhu2020msnet} which includes user-generated aerial videos collected from social media platform.

\par Besides satellite imagery, very few UAV based disaster datasets are also available. Kyrkou \textit{et al.} proposed an image classification dataset named AIDER (Aerial Image Database for Emergency Response) \cite{kyrkou2019deep}. AIDER consists of images from four different disaster events including Fire/Smoke, Flood, Collapsed Building/Rubble, and Traffic Accidents. Rahnemoonfar \textit{et al.} presented a high resolution post-hurricane dataset named FloodNet \cite{rahnemoonfar2021floodnet}. This dataset provides pixel level annotation of nine classes for natural disaster damage assessment. The UAV images provided by FloodNet consists of flooded and non-flooded areas collected after hurricane Harvey.

\par
\begin{figure*}[t]
	\begin{center}
		\includegraphics[width=\linewidth]{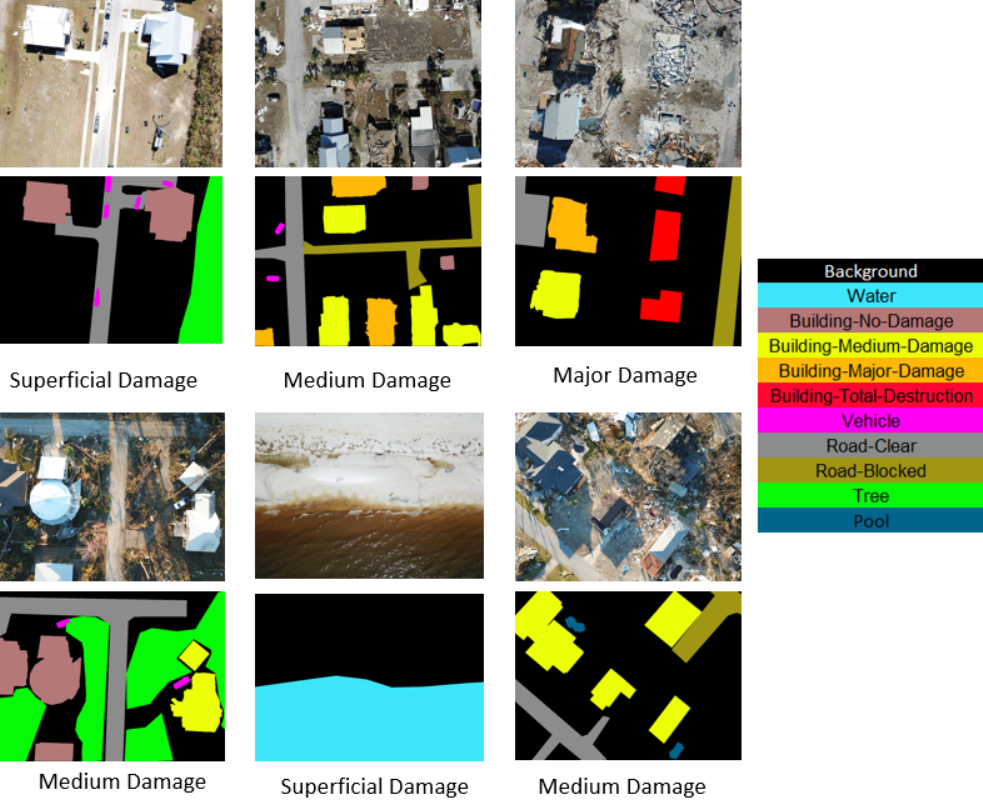}
	\end{center}
	\caption{Illustration of complex scenes of RescueNet dataset. First and third rows show original images, and the respective lower rows show the corresponding annotations (both semantic segmentation and image classification).}
	\label{fig:rescuenet-example-img}
\end{figure*}

 In the field of natural disaster damage assessment, the research community currently faces two significant challenges. Firstly, there is a lack of comprehensive pixel-level annotation for post-disaster scenes. This issue arises due to the absence of datasets that provide thorough annotation of the entire scene at the pixel level. The few available datasets only offer annotation for a limited number of classes \cite{rudner2019multi3net, gupta2019creating, zhu2020msnet}.
However, damage information for only a few classes does not provide a complete understanding of the scene. Other elements present in the images, such as roads, vehicles, damaged trees, and debris, play crucial roles in explaining the scene comprehensively and aiding in more accurate damage assessment. To address this gap, this paper introduces RescueNet, a low altitude and high-resolution natural disaster dataset. RescueNet provides pixel-level annotation for 10 classes, expanding across six distinct categories, including water, buildings, vehicles, roads, trees, and pools.
The second challenge pertains to the classification of different damage levels within a category. For instance, after a hurricane or wildfire, a building or road can be damaged, but the extent of the damage is often not specified in most datasets. Moreover, a building can sustain different levels of damage, ranging from slight damage to complete destruction after a disaster. While xBD dataset \cite{gupta2019creating} classifies buildings based on four different damage levels, it is a low-resolution satellite dataset, resulting in poor image quality . To address this limitation, RescueNet provides building segmentation labels with four distinct damage criteria, while ensuring the images are of very high resolution. Figure \ref{fig:rescuenet-example-img} illustrates some examples from the RescueNet dataset, showcasing the complexity of the scenes along with their corresponding annotations. Our annotation includes both classification and semantic segmentation labels on high resolution imagery. In addition to offering semantic segmentation annotation for different categories and sub-categories such as buildings and roads, RescueNet also provides three-class classifications for each neighborhood. These classifications include superficial damage, medium damage, and major damage. An initial version of RescueNet was published as HRUD \cite{chowdhury2020comprehensive} where only 1973 images were released. There are primarily four distinctions between HRUD and RescueNet: 1) the number of images has been increased from 1973 to 4494 in RescueNet, 2) the object class ``Debris'' and ``Sand'' (present in the HRUD) have been combined with the background class in RescueNet, 3) ``Road'' class has been divided into two classes namely ``Road-Clear'' and ``Road-Blocked'' in RescueNet, and 4) three image label neighborhood damage classifications have been introduced in RescueNet. A comparative study of various existing natural disaster datasets with RescueNet is presented in Table \ref{table:overview-realted-works}. As depicted in Table \ref{table:overview-realted-works}, it is evident that FloodNet and RescueNet are the sole datasets that offer comprehensive semantic segmentation labels for high-resolution UAV imagery.

In summary, the contributions of this paper are as follows:

\begin{itemize}
 \item Introduction of RescueNet, a high-resolution aerial imagery dataset captured using unmanned aerial vehicles (UAVs).

 \item Provision of high-quality annotation for 10 classes, including images from both affected and non-affected areas following Hurricane Michael.

 \item Annotation of building and road damages based on severity. Building damages are classified into four damage classes: superficial damage, medium damage, major damage, and total destruction. Roads are classified as either road-clear or road-blocked.

 \item Evaluation of various existing semantic segmentation methods on the RescueNet dataset, demonstrating its utility in future research on natural disaster damage assessment.

\end{itemize}

\begin{table*}[!htp]
	\centering
	\caption{Overview of existing natural disaster datasets.}\label{table:overview-realted-works}
	\scalebox{1}
	{
	\begin{tabular}{l c c c c c}
	\hline
		Dataset & Size & Resolution & Image Type & Task & \begin{tabular}{@{}c@{}}\# of Annotated \\  Classes\end{tabular} \\ \hline
		ABCD \cite{fujita2017damage} & 22171 & Varies & Satellite & Classification & 2 \\ \hline
		Chen \textit{et al.} \cite{chen2018benchmark} & - & Varies & Satellite & Object Detection & 2 \\ \hline
		fMoW \cite{christie2018functional} & $\sim$ 1 million & Varies & Satellite & Classification & 63 \\ \hline
		xBD \cite{gupta2019creating} & 22068 & 1024x1024 & Satellite & \begin{tabular}{@{}c@{}}Instance Segmentation, \\ Classification\end{tabular} & 4 \\ \hline
		ISBDA \cite{zhu2020msnet} & 1030 & - & \begin{tabular}{@{}c@{}}Aerial \\  (Social Media)\end{tabular}  & Object Detection & 3 \\ \hline
		AIDER \cite{kyrkou2019deep} & 2545 & - & UAV & Classification & 4 \\ \hline		
		FloodNet \cite{rahnemoonfar2021floodnet} & 2343 & 3000x4000 & UAV & \begin{tabular}{@{}c@{}}Semantic Segmentation,\\  Classification\end{tabular} & 9 \\ \hline
		\textbf{RescueNet} & \textbf{4494} & \textbf{3000x4000} & \textbf{UAV} & \textbf{\begin{tabular}{@{}c@{}}Semantic Segmentation,\\ Classification\end{tabular}} & \textbf{10} \\ \hline
	\end{tabular}
	}
\end{table*}

\section*{Methods}

\subsection*{Data Collection} \label{subsection:data-collection}

\par Hurricane Michael made landfall near Mexico Beach, Florida, on October 10, 2018, as a category 5 hurricane, one of the powerful and destructive tropical cyclones to strike United States since Andrew in 1992. The dataset was collected by the Center for Robot-Assisted Search and Rescue using small unmanned aerial vehicles (sUAVs) on behalf of Florida State Emergency Response Team at Mexico Beach and other directly impacted areas \cite{michael} after operating 80 flights conducted between October 11-14, 2018. The two most important features of this dataset are fidelity and uniqueness. Firstly, the data is collected by emergency responders during the response phase, utilizing small unmanned aerial vehicles (sUAVs). This ensures that the dataset aligns with current data collection practices and represents the information typically gathered during a disaster. Secondly, it is unique since it is the only known database of sUAV imagery for disaster. Note that there is other existing database of imagery collected using unmanned and manned aerial assets during different disasters, such as National Guard Predators or Civil Air Patrol. But compared to our dataset, those are collected using larger and fixed-wing assets that operate above 400 feet AGL (above ground level). While images were taken using DJI Mavic Pro quadcopters, two sets of videos were taken with Parrot Disco fixed-wing sUAV and one set at night with a DJI Inspire and thermal camera. But these videos are not included in the RescueNet.

\begin{figure*}[t]
	\begin{center}
		\includegraphics[width=\linewidth]{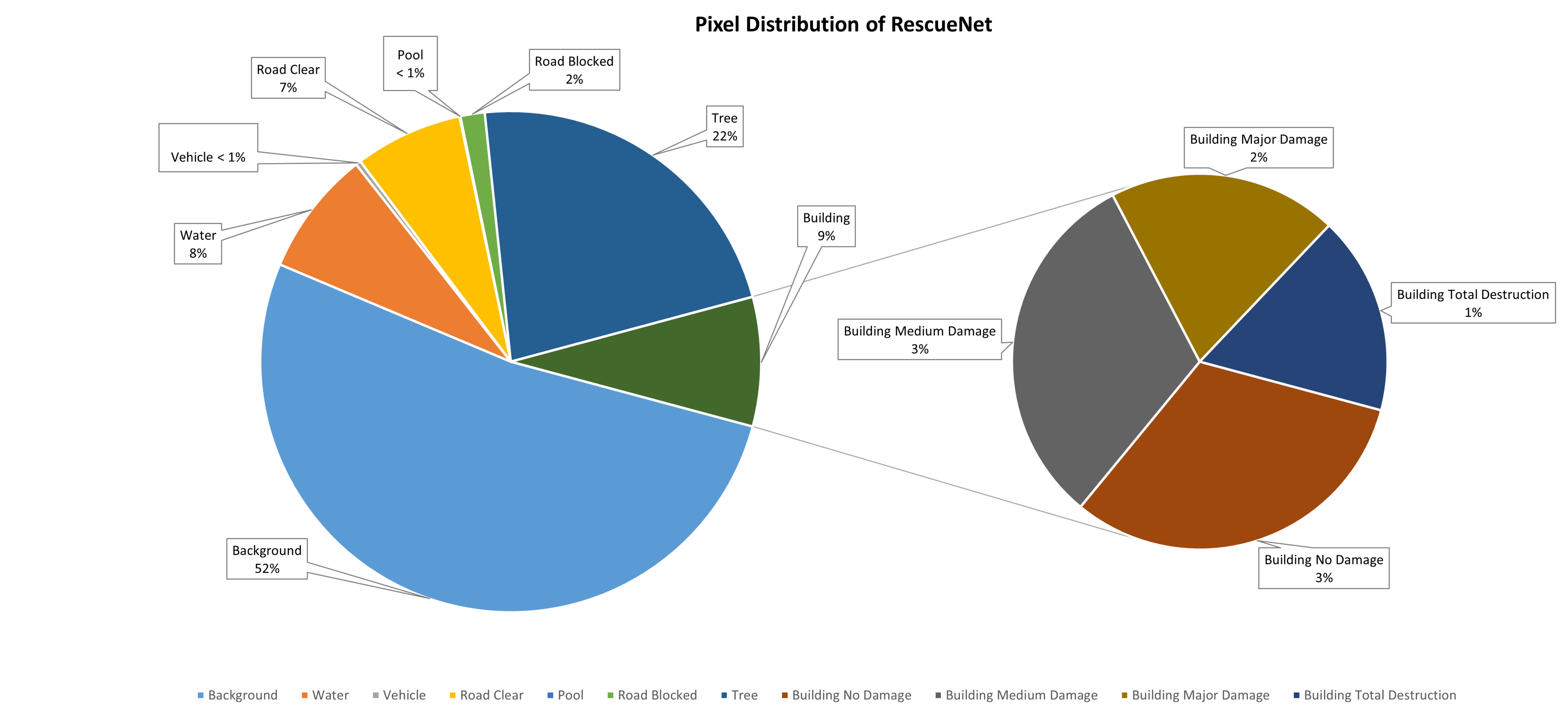}
	\end{center}
	\caption{Pixel distribution of different classes in RescueNet.}
	\label{fig:pie-plot-pixels}
\end{figure*}

\subsection*{Data Annotation} \label{subsection:data-annotation}

\subsubsection*{Annotation for Semantic Segmentation}
\par The V7 Darwin platform (V7Darwin) \cite{V7Darwin} is employed for the annotation of the dataset for semantic segmentation. The objective is to achieve comprehensive pixel-level annotation for each image in the dataset. To accomplish this, all objects present in the dataset are meticulously annotated. This allows us to gain a comprehensive understanding of the extent of damage caused by natural disasters. The objects annotated in the datasets are buildings, roads, trees, water, vehicles, and pools. The building class encompasses both residential and non-residential structures. In accordance with the FEMA guideline \cite{fema2020fema}, the building damages are further categorized into four levels: Building-No-Damage, Building-Medium-Damage, Building-Major-Damage, and Building-Total-Destruction. Table \ref{table:building-stat} provides a summary of the annotated polygons for these four classes.

\begin{table}[!htp]
	\centering
	\caption{Number of polygons of different buildings based on their damage levels.}\label{table:building-stat}
	\begin{tabular}{l c}
	\hline
		Damage Level & Number of Polygons\\\hline
		\hline
		No Damage & 4011 \\
		Medium Damage & 3119\\
		Major Damage & 1693\\
		Total Destruction & 2080\\ \hline
	\end{tabular}
\end{table}

\par Although the FEMA guideline \cite{fema2020fema} includes scenarios where the conditions of buildings can be observed from all sides, aerial images provide only top views. Therefore, it becomes necessary to adapt the definitions based on these top views. Based on input from first responders, the building damage classes are defined as follows:
\begin{itemize}

\item Building-No-Damage: This class is assigned when no damage is observed on a building.

\item Building-Medium-Damage: This class is assigned when certain parts of a building are affected, requiring minimal repairs to make it habitable. For example, if the roof can be temporarily covered with a blue tarp, it falls under this classification.

\item Building-Major-Damage: This class is assigned when the damage sustained by a building is significant enough to require extensive repairs, indicating substantial structural damage.

\item Building-Total-Destruction: This class is assigned when two or more major structural components of a building collapse, such as basement walls, foundation, load-bearing walls, or the roof.

\end{itemize}
It is important to note that these definitions are adapted specifically for the top-view perspective provided by aerial images, taking into consideration the information and observations from first responders. 

\par The road annotations in RescueNet dataset are divided into two classes: Road Clear and Road Blocked. In the aftermath of a natural disaster, roads can become obstructed by various obstacles, including floodwater, sand, and debris from trees and buildings. These obstructed roads are challenging to access and are classified as "Road Blocked" in the dataset. On the other hand, roads that are not covered by any obstacles are classified as "Road Clear". The defining criteria for all classes, including buildings and roads, are summarized in Table \ref{table:dataset-class-def}.

\par Several examples from the RescueNet dataset, along with their corresponding colored annotated masks, are depicted in Figure \ref{fig:rescuenet-example-img}. In the first image, various undamaged buildings, vehicles, trees, and roads can be observed. The second image displays buildings with different levels of damage, alongside roads. Notably, the right side of the road is obstructed by debris, leading to its annotation as "Road Blocked," while the left side is annotated as "Road Clear." The third image showcases another damaged area with buildings classified into different damage categories, as well as a road. Due to the presence of debris, the road is annotated as "Road Blocked." Similarly, three more examples are presented in the third row, alongside their corresponding annotations displayed in the fourth row.

\subsubsection*{Annotation for Image Classification}
\par Thus far, the discussion has primarily focused on the annotation process for semantic segmentation. However, in addition to pixel-level annotation, each image in the dataset is also classified into one of three classes: Superficial damage, Medium damage, and Major damage. This classification serves to represent the overall damage level of a specific area captured by the image. It serves as an image classification tag for each image.

The classification of an image is based on the extent of damage within the area covered by that image, encompassing both man-made and natural structures. If the image does not exhibit any damaged structures, it is classified as "Superficial damage". In the case where a few structures are damaged by the natural disaster, the image is labeled as "Medium damage". Finally, if the image contains at least one completely destroyed building or if approximately 50\% of the area is covered with debris, it is categorized as "Major damage". 

\subsection*{Quality Control} \label{subsection:quality-control}

\par To ensure the quality of the annotation, a rigorous two-step verification process is employed for each image. An annotation guideline, which includes the definition of different classes and distinguishing features for various building damages as described in the "Data Annotation" section, is provided to the annotators. The annotation process involves pixel-level annotation of all objects present in the image, making it a time-consuming task that takes approximately one hour per image.

\par Quality control measures are implemented to maintain the accuracy and consistency of the annotations. Initially, each image is assigned to an annotator who performs the annotation. Once the annotation is completed, the image is passed on to a reviewer for verification. The reviewer meticulously examines the quality of the annotation, ensuring its adherence to the guidelines and overall accuracy. If any discrepancies or inaccuracies are identified, the image is returned to the annotator with detailed comments for further correction. This iterative cycle of review and correction continues until the reviewers are satisfied with the quality of the annotations. Through this meticulous review process, the dataset attains a high standard of annotation quality and consistency.

\begin{table}[ht]
\caption{Instance types and classes in RescueNet.}
\begin{center}
\begin{adjustbox}{width=\linewidth}
\begin{tabular}{c|c|l}
    \hline 
    Instance Type & Classes & \multicolumn{1}{c}{Definition} \\
    \hline \hline
    Water & Water & This class encompasses collections of water that are identified as natural water reservoirs or caused by floods. \\
    \hline
     \multirow{4}{*}{Building} & Building No Damage & Houses or structures that exhibit no or minimal visible damage on their roofs are classified as "Building No Damage". \\
     
    & Building Minor Damage & Buildings with partially damaged roofs, allowing the use of tarps for coverage, fall under the category of "Building Minor Damage". \\
    
    & Building Major Damage &  Buildings with severely damaged roofs, either completely or substantially, are classified as "Building Major Damage"'.\\
    
    & Building Total Destruction & Buildings that are completely destroyed, only the base visible, are categorized as "Building Total Destruction".\\
    
    \hline
    Vehicle & Vehicle & This class includes various types of transportation vehicles, such as trucks and cars.\\
    \hline
    \multirow{2}{*}{Road} & Road-Clear & Roads that are undamaged and free from debris or water are classified as "Road Clear". \\
    & Road-Blocked & Roads that are obstructed by debris or flooded are categorized as "Road Blocked".\\
    \hline
    Tree & Tree & This class encompasses individual trees or groups of trees, regardless of whether they are damaged or undamaged. \\
    \hline
    Pool & Pool & The presence of a pool, regardless of whether it contains water, located adjacent to a building, is classified as "Pool".\\
    \hline \hline
\end{tabular}
\end{adjustbox}
\end{center}
\label{table:dataset-class-def}
\end{table}

\subsection*{Generation of Segmentation Mask and Classification Label} \label{subsection:mask-generation}

\par Annotation are performed on V7 Darwin Platform \cite{V7Darwin}. Annotation of each image are recorded in ``json'' data format. In the json files, objects of different classes are recorded as \textit{bounding boxes} in dictionary format (key-value pair). Dictionary description of each object has several keys including \textit{polygons}, \textit{name}, and \textit{attributes}. Different keys have different functionalities. \textit{Polygon} acts as a bounding box of an object and contains corner points of that polygon. Each pixel inside of a polygon represents the corresponding class. The \textit{name} denotes the class name of an object and the \textit{attributes} key represents the sub-classification of that object class (if there is any). For example, if \textit{name} denotes a ``building'' class, then \textit{attributes} can denotes any of the four subclasses such as ``No (Superficial) Damage'', ``Minor (Medium) Damage'', ``Major Damage'', and ``Total Destruction''. In addition each json file contains a \textit{tag} (another type of dictionary entry) which has a \textit{name} key whose value starts with term ``Neighborhood'' to represent the overall damage classification of the image. For example, depending on the damage amount observed in an image, the \textit{name} key can have one of the three classes: ``Superficial Damage'', ``Medium Damage'', and ``Major Damage''.

\par To generate masks for semantic segmentation task we utilize the information recorded in \textit{bounding box} components of json files. As mentioned in the previous paragraph each json file contains \textit{bounding box} dictionaries which keep records of object classification of each pixel. For an image we start with an empty mask and then update each pixel value by going through each \textit{bounding box} of the corresponding json file. Following this procedure for each image, segmentation masks are generated for the entire dataset. The pixels with ``Water'' classes are labeled as 1, ``Building No Damage'' as 2, ``Building Minor Damage'' as 3, ``Building Major Damage'' as 4, ``Building Total Destruction'' as 5, ``Vehicle'' as 6, ``Road-Clear'' as 7, ``Road-Blocked'' as 8, ``Tree'' as 9, and ``Pool'' as 10. Rest of the pixels are considered ``Background'' and labeled as 0. On the other hand, to generate classification labels for image classification task we extract labels from the ``Neighborhood'' \textit{tag} of the json file. The classification labels of images are recorded in ``.csv'' files where classes are represented as 0 when the classification label is ``Superficial Damage'', 1 when classification label is ``Medium Damage'', and 2 when the label is ``Major Damage''.

\section*{Data Records}

\subsection*{Dataset Statistics} \label{subsection:dataset-statistics}

\subsubsection*{Semantic Segmentation}
RescueNet \cite{rahnemoonfar:rescuenet2023} is a dataset collected from the areas affected by Hurricane Michael, utilizing UAVs. The dataset comprises multiple classes, including water, building-no-damage, building-minor-damage, building-major-damage, building-total-destruction, vehicle, road-clear, road-blocked, tree, and pool. Notably, RescueNet offers an extensive range of semantic labels specifically for buildings, based on their respective damage levels. Table \ref{table:building-stat} provides an overview of the dataset's building annotations, indicating the presence of 4,011 polygons labeled as building-no-damage, 3,119 polygons labeled as building-minor-damage, 1,693 polygons labeled as building-major-damage, and 2,080 polygons labeled as building-total-destruction. The distribution of pixels across different classes in the RescueNet dataset is visualized in Figure \ref{fig:pie-plot-pixels}. These statistics and visualizations demonstrate the comprehensive coverage and diversity of the dataset, particularly in terms of building damage annotations. Some sample images from this category are shown in Figure \ref{fig:rescuenet-example-img}.

\subsubsection*{Image Classification}
In addition to pixel-level annotation (semantic segmentation), the RescueNet \cite{rahnemoonfar:rescuenet2023} dataset also includes image-level annotations. Each image in the dataset is categorized into one of three damage levels: superficial damage, medium damage, or major damage. In RescueNet, 1389 images are classified as ``Superficial Damage'', 1906 images as ``Medium Damage'', and 1199 images as ``Major Damage''. Some sample images from this category are shown in Figure \ref{fig:rescuenet-example-img}.

\begin{table}[ht]
\caption{Distribution of classification labels (Superficial Damage, Medium Damage, Major Damage) in RescueNet.}
\begin{center}

\begin{tabular}{c|ccc|c}
    \hline 
    Classification Type & Train & Validation & Test & Total\\
    \hline \hline
    Superficial Damage & 1111 & 139 & 139 & 1389 \\
    \hline
    Medium Damage & 1525 & 190 & 191 & 1906\\
    \hline
    Major Damage & 959 & 120 & 120 & 1199\\
    \hline \hline
\end{tabular}
\end{center}
\label{table:stats-classification}
\end{table}

\subsection*{Dataset Splits} \label{subsection:dataset-splits}

\par We partitioned our dataset into three subsets: training, validation, and testing. However, it should be noted that the dataset includes both areas affected by disasters and non-affected areas. Some images depict areas covered in debris with a variety of buildings exhibiting different damage labels, such as superficial, medium, major, or total destruction. To ensure a balanced distribution of images from both affected and non-affected areas across the three subsets, we utilized the image classification of the dataset where all images of the dataset are categorized into one of the three categories: superficial damage, medium damage, and major damage.

\par From each category, 80\% of the images were allocated to the training set, 10\% to the validation set, and the remaining 10\% to the test set. This distribution strategy ensured a representative distribution of images across the subsets. The training set consists of 3,595 images, the validation set contains 449 images, and the test set contains 450 images. The distribution of image classes (Superficial, Medium, Major) in the dataset is shown in Table \ref{table:stats-classification}.

\section*{Technical Validation}

\subsection*{Problem Details}
\par The objective of semantic segmentation, one of the fundamental tasks in computer vision, is to assign label to each pixel of an image. RescueNet \cite{rahnemoonfar:rescuenet2023} consists of post-disaster images and provide pixel level annotation of 10 object classes. For any image in the dataset, the RescueNet has annotation of all objects present in the scene along with detailed damage levels of some of the classes such as buildings, and roads.

\subsection*{Models and Training Details}
\subsubsection*{Selected Semantic Segmentation Models}
\par We implement four state-of-the-art semantic segmentation methods namely PSPNet \cite{zhao2017pyramid}, DeepLabv3+ \cite{chen2018encoder}, Segmenter \cite{strudel2021segmenter}, Attention UNet \cite{oktay2018attention}, and evaluated their performance on RescueNet.  


\par PSPNet architecture can be divided into two stages: feature map generation and pyramid pooling module. The first stage generate feature map from an input image either using transfer learning or from scratch using dilated convolutions. Dilated convolutions gather large size area information using smaller kernels for higher dilation rates by keeping dimension same as input image. In the pyramid pooling module, feature maps are average pooled at different pool size. The reason behind using different pooling scales is to correctly segment objects of all sizes. Since an image contains objects of various sizes ranging from small area to large area in different regions, pooling at different scales can segment objects with different sizes. DeepLabv3+ \cite{chen2018encoder} is an encoder-decoder semantic segmentation model which upgrades DeepLabv3 \cite{chen2017rethinking} by solving the issue of lowering of prediction accuracy and loss of boundary information due to multiple downsampling operations. DeepLabv3+ implements atrous convolution/dilated convolution to enlarge the field of view of the kernels and thus solving the issue of usage of regular downsampling operations. Moreover, application of  depthwise separable convolution to both atrous spatial pyramid pooling and decoder modules results in faster segmentation network. Attention U-Net \cite{oktay2018attention} introduces a novel grid-based attention gate module which allows attention coefficients to be more specific to local regions. This gating mechanism has been added to U-Net \cite{ronneberger2015u} as an extension which improves the model sensitivity to foreground pixels without  requiring complicated heuristics. Finally, Segmenter \cite{strudel2021segmenter} is a vision transformer based segmentation network which relies on the output embeddings corresponding
to image patches and obtain class labels from these embeddings with a point-wise linear decoder or a mask transformer decoder.

\subsubsection*{Training Details}
\par We implement the methods using PyTorch and use NVIDIA GeForce RTX 2080 Ti GPU and Intel Core i9 CPU as hardware. We use resnet101 as backbone for PSPNet and DeepLabv3+. We implement ``poly'' learning rate where base learning rate is 0.001. All the models use the following hyperparameters settings. Momentum is set to 0.9, weight decay to 0.00001, power to 0.9, and weight of the auxiliary rate to 0.4. For data augmentation we implement scaling, flipping, random shuffling, and random rotation. Data augmentation helps in avoiding overfitting. We resize the images to 713 $\times$ 713 during training. 


\begin{table*}[!htp]
	\centering
	\caption{Per-class results on RescueNet testing set.}\label{table:rescuenet-model-test-perform-table}
	\begin{adjustbox}{width=\linewidth}
	\begin{tabular}{c | c c c c c c c c c c c c c| c}
	\hline
		Method & Water &  \begin{tabular}{@{}c@{}}Building  \\  No Damage\end{tabular} & \begin{tabular}{@{}c@{}}Building  \\  Minor Damage\end{tabular} & \begin{tabular}{@{}c@{}}Building  \\  Major Damage\end{tabular} & \begin{tabular}{@{}c@{}}Building \\  Total Destruction\end{tabular} & Vehicle & Road-Clear & Road-Blocked & Tree & Pool & \textbf{Mean IoU \%}\\\hline
		\hline
            DeepLabv3+\cite{chen2018encoder} & 78.4 & 61.6 & 49.6 & 47.3 & 57.2 & 49.6 & 69.7 & 32.9 & 78.6 & 49.4 & 57.43\\ \hline
            \begin{tabular}{@{}c@{}}Segmenter\cite{strudel2021segmenter}  \\  (ViT-Tiny)\end{tabular} & 94.77 & 65.18 & 44.41 & 53.16 & 89.54 & 47.70 & 87.79 & 51.14 & 97.17 & 54.00 & 68.49\\ \hline
            \begin{tabular}{@{}c@{}}Segmenter\cite{strudel2021segmenter}  \\  (ViT-Small)\end{tabular} & 97.46 & 80.74 & 72.42 & 79.67 & 95.85 & 66.69 & 94.63 & 84.10 & 98.58 & 87.66 & 85.78\\ \hline
	    PSPNet\cite{zhao2017pyramid} & 98.74 & 95.16 & 94.36 & 96.91 & 98.95 & 85.97 & 98.07 & 94.83 & 99.40 & 94.31 & 95.67\\ \hline
		Attention UNet \cite{oktay2018attention} & 99.15 & 99.46 & 99.59 & 99.56 & 99.89 & 91.16 & 98.94 & 98.72 & 99.63 & 98.62 & \bftab 98.47\\ \hline \hline
	\end{tabular}
	\end{adjustbox}
\end{table*}

\begin{figure*}[t]
	\begin{center}
		\includegraphics[width=\linewidth]{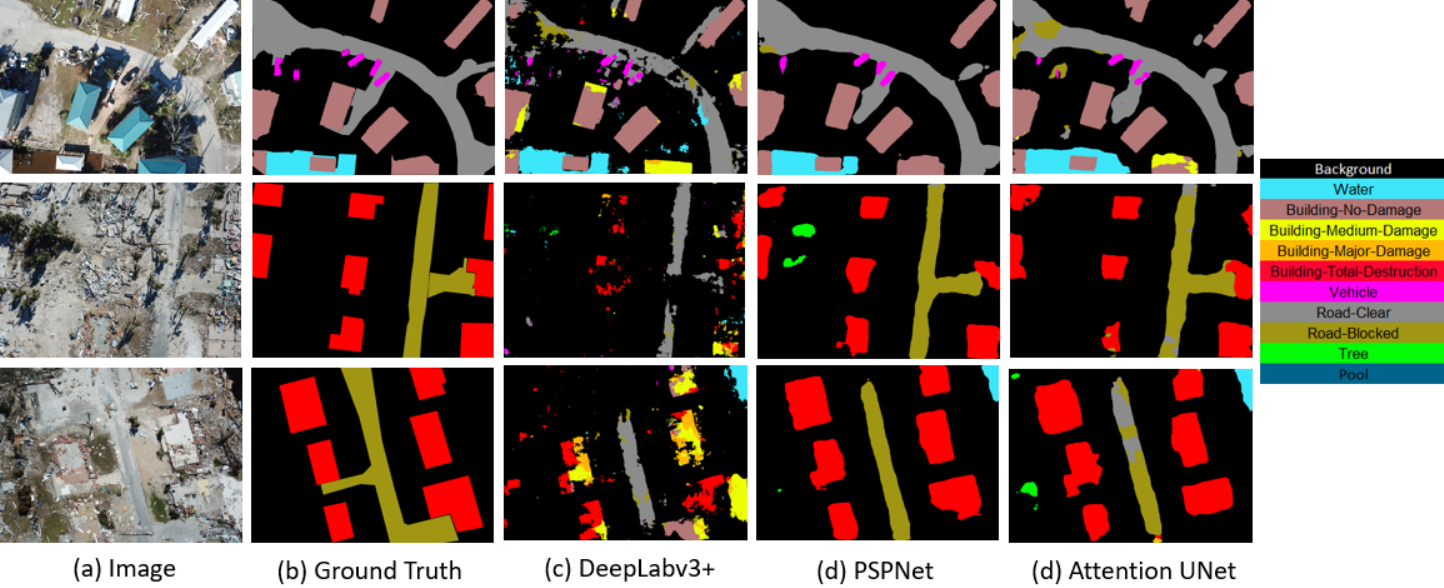}
	\end{center}
	\caption{Visual comparison of semantic segmentation algorithms on RescueNet test set.}
	\label{fig:vis-compare-models-all-cls-rescuenetv1.0}
\end{figure*}

\begin{table*}[!htp]
	\centering
	\caption{Per-class results on FloodNet testing set based on RescueNet pretrained weights ($\checkmark$ represents transfer leaning based training and \xmark \; represents non-transfer learning based training).}\label{table:floodnet-segmentation-test-perfor-table}
	\begin{adjustbox}{width=\textwidth}
	\begin{tabular}{l c c c c c c c c c c| c}
	\hline
		Method & \begin{tabular}{@{}c@{}}Pretrained on \\  RescueNet\end{tabular} & Building Flooded & \begin{tabular}{@{}c@{}}Building Non \\  Flooded\end{tabular} & Road Flooded &  \begin{tabular}{@{}c@{}}Road Non \\  Flooded \end{tabular} & Water & Tree & Vehicle & Pool & Grass & mIoU\\\hline
		\hline
		DeepLabv3+\cite{chen2018encoder} & \xmark & 28.10 & 78.10 & 32.00 & 81.10 & 73.00 & 74.50 & 33.60 & 40.00 & 87.10 & 58.61\\
            DeepLabv3+\cite{chen2018encoder} & \checkmark & 49.3 & 77.7 & 43.2 & 81.00 & 69.00 & 77.9 & 39.5 & 44.8 & 86.1 & \bftab 63.16\\  \hline
		PSPNet\cite{zhao2017pyramid} & \xmark & 65.61 & 90.92 & 78.69 & 90.90 & 91.25 & 89.17 & 54.83 & 66.37 & 95.45 & 80.35\\
            PSPNet\cite{zhao2017pyramid} & \checkmark & 81.80 & 91.37 & 79.30 & 90.03 & 90.14 & 87.90 & 66.43 & 74.48 & 94.52 & \bftab 83.99\\ \hline
	\end{tabular}
	\end{adjustbox}
\end{table*}

\subsection*{Evaluation Metric}
\par There are several evaluation metrics for computer vision like Accuracy, F1 Score, and  Intersection Over Union (IoU). However, IoU is the most popular and precise metric for the evaluation of the performance of a segmentation model. IoU metric, also referred to as the Jaccard index, is essentially a method to quantify the percent overlap between the target mask and the predicted output. The IoU metric measures the number of pixels common between the target mask and predicted mask divided by the total number of pixels present across both masks.

\begin{equation*}
    IoU = \frac{target \; \bigcap \; prediction}{target \; \bigcup \; prediction}
\end{equation*}

\subsection*{Experimental Result Analysis}

\par The experimental results are presented in Table \ref{table:rescuenet-model-test-perform-table}. It is evident from the table that the attention-based method, Attention UNet \cite{oktay2018attention}, achieves the best performance compared to the non-attention-based methods. Among the non-attention-based methods, PSPNet \cite{zhao2017pyramid}, which implements a pyramid pooling architecture, demonstrates better performance than the atrous convolutional operation-based method DeepLabv3+ \cite{chen2018encoder}. Performance of a transformer based method Segmenter have also been presented in Table \ref{table:rescuenet-model-test-perform-table} with two different backbones: ViT-Tiny and ViT-Small. From the experimental results it is evident transformer based model with heavier backbone achieves better results. All these results indicate that the proposed RescueNet dataset can be effectively utilized to extract scene information and detect objects after a disaster using deep learning methods.

\par RescueNet has also been explored from the perspective of transfer learning. We performed transfer learning on a similar semantic segmentation dataset named FloodNet by implementing DeepLabv3+ and PSPNet. In transfer learning framework, we first train the deep learning models on RescueNet and then use the learned weights to train on FloodNet. The results are shown in Table \ref{table:floodnet-segmentation-test-perfor-table}. From the results we can observe that transfer learning based training improves the performance of deep learning models on segmentation task. Therefore, besides sole usage of RescueNet, it can also be utilized to improve the accuracy of deep learning models on other datasets.
    
\section*{Usage Notes}

There are several important practical and academic research applications of RescueNet:

\par \textbf{Damage classification of buildings:} RescueNet provides explicit classification of building damages based on different damage levels. This information is crucial for first responders and rescue planners, as it allows them to make accurate decisions regarding the allocation of rescue efforts.

\par \textbf{Road segmentation with debris:} RescueNet includes segmented images of roads, classified as either "Road-Clear" or "Road-Blocked". The label "Road-Blocked" indicates that the roads are covered with flood water or debris from destroyed buildings or trees. Conversely, the label "Road-Clear" indicates that the roads are undamaged and not obstructed by any obstacles. A machine learning model with accurate segmentation of both categories of roads can greatly improve rescue planning and assist the rescue team in finding the most efficient routes to reach those affected by the disaster.

\par \textbf{Usability in future disaster events:} Given the occurrence of various natural disasters worldwide every year, it is crucial to accurately estimate the damage caused by these events. Understanding the different components of each affected area is imperative in this process. By training a machine learning model with RescueNet, it becomes possible to promptly identify damages in these areas, which can help save human lives and reduce infrastructure costs. Furthermore, RescueNet can be combined with other datasets and utilized in semi-supervised and self-supervised learning approaches to leverage its rich semantic labels, thereby enhancing the effectiveness of disaster assessment and response strategies.

\section*{Code availability}

\par To reproduce the semantic segmentation results, the source code are available at \url{https://github.com/BinaLab/RescueNet-A-High-Resolution-Post-Disaster-UAV-Dataset-for-Semantic-Segmentation/tree/main}.

\bibliography{00-main}

\begin{thebibliography}{10}
\urlstyle{rm}
\expandafter\ifx\csname url\endcsname\relax
  \def\url#1{\texttt{#1}}\fi
\expandafter\ifx\csname urlprefix\endcsname\relax\def\urlprefix{URL }\fi
\expandafter\ifx\csname doiprefix\endcsname\relax\def\doiprefix{DOI: }\fi
\providecommand{\bibinfo}[2]{#2}
\providecommand{\eprint}[2][]{\url{#2}}

\bibitem{cordts2016cityscapes}
\bibinfo{author}{Cordts, M.} \emph{et~al.}
\newblock \bibinfo{title}{The cityscapes dataset for semantic urban scene understanding}.
\newblock In \emph{\bibinfo{booktitle}{Proceedings of the IEEE conference on computer vision and pattern recognition}}, \bibinfo{pages}{3213--3223} (\bibinfo{year}{2016}).

\bibitem{everingham2010pascal}
\bibinfo{author}{Everingham, M.}, \bibinfo{author}{Van~Gool, L.}, \bibinfo{author}{Williams, C.~K.}, \bibinfo{author}{Winn, J.} \& \bibinfo{author}{Zisserman, A.}
\newblock \bibinfo{journal}{\bibinfo{title}{The pascal visual object classes (voc) challenge}}.
\newblock {\emph{\JournalTitle{International journal of computer vision}}} \textbf{\bibinfo{volume}{88}}, \bibinfo{pages}{303--338} (\bibinfo{year}{2010}).

\bibitem{mottaghi2014role}
\bibinfo{author}{Mottaghi, R.} \emph{et~al.}
\newblock \bibinfo{title}{The role of context for object detection and semantic segmentation in the wild}.
\newblock In \emph{\bibinfo{booktitle}{Proceedings of the IEEE Conference on Computer Vision and Pattern Recognition}}, \bibinfo{pages}{891--898} (\bibinfo{year}{2014}).

\bibitem{caesar2018coco}
\bibinfo{author}{Caesar, H.}, \bibinfo{author}{Uijlings, J.} \& \bibinfo{author}{Ferrari, V.}
\newblock \bibinfo{title}{Coco-stuff: Thing and stuff classes in context}.
\newblock In \emph{\bibinfo{booktitle}{Proceedings of the IEEE conference on computer vision and pattern recognition}}, \bibinfo{pages}{1209--1218} (\bibinfo{year}{2018}).

\bibitem{nguyen2017damage}
\bibinfo{author}{Nguyen, D.~T.}, \bibinfo{author}{Ofli, F.}, \bibinfo{author}{Imran, M.} \& \bibinfo{author}{Mitra, P.}
\newblock \bibinfo{title}{Damage assessment from social media imagery data during disasters}.
\newblock In \emph{\bibinfo{booktitle}{Proceedings of the 2017 IEEE/ACM International Conference on Advances in Social Networks Analysis and Mining 2017}}, \bibinfo{pages}{569--576} (\bibinfo{year}{2017}).

\bibitem{nia2017building}
\bibinfo{author}{Nia, K.~R.} \& \bibinfo{author}{Mori, G.}
\newblock \bibinfo{title}{Building damage assessment using deep learning and ground-level image data}.
\newblock In \emph{\bibinfo{booktitle}{2017 14th conference on computer and robot vision (CRV)}}, \bibinfo{pages}{95--102} (\bibinfo{organization}{IEEE}, \bibinfo{year}{2017}).

\bibitem{weber2020detecting}
\bibinfo{author}{Weber, E.} \emph{et~al.}
\newblock \bibinfo{title}{Detecting natural disasters, damage, and incidents in the wild}.
\newblock In \emph{\bibinfo{booktitle}{European Conference on Computer Vision}}, \bibinfo{pages}{331--350} (\bibinfo{organization}{Springer}, \bibinfo{year}{2020}).

\bibitem{chen2018benchmark}
\bibinfo{author}{Chen, S.~A.} \emph{et~al.}
\newblock \bibinfo{journal}{\bibinfo{title}{Benchmark dataset for automatic damaged building detection from post-hurricane remotely sensed imagery}}.
\newblock {\emph{\JournalTitle{arXiv preprint arXiv:1812.05581}}}  (\bibinfo{year}{2018}).

\bibitem{gupta2019creating}
\bibinfo{author}{Gupta, R.} \emph{et~al.}
\newblock \bibinfo{title}{Creating xbd: A dataset for assessing building damage from satellite imagery}.
\newblock In \emph{\bibinfo{booktitle}{Proceedings of the IEEE Conference on Computer Vision and Pattern Recognition Workshops}}, \bibinfo{pages}{10--17} (\bibinfo{year}{2019}).

\bibitem{rahnemoonfar2021floodnet}
\bibinfo{author}{Rahnemoonfar, M.} \emph{et~al.}
\newblock \bibinfo{journal}{\bibinfo{title}{Floodnet: A high resolution aerial imagery dataset for post flood scene understanding}}.
\newblock {\emph{\JournalTitle{IEEE Access}}} \textbf{\bibinfo{volume}{9}}, \bibinfo{pages}{89644--89654} (\bibinfo{year}{2021}).

\bibitem{nguyen2017automatic}
\bibinfo{author}{Nguyen, D.~T.}, \bibinfo{author}{Alam, F.}, \bibinfo{author}{Ofli, F.} \& \bibinfo{author}{Imran, M.}
\newblock \bibinfo{journal}{\bibinfo{title}{Automatic image filtering on social networks using deep learning and perceptual hashing during crises}}.
\newblock {\emph{\JournalTitle{arXiv preprint arXiv:1704.02602}}}  (\bibinfo{year}{2017}).

\bibitem{weber2020eccv}
\bibinfo{author}{Weber, E.} \emph{et~al.}
\newblock \bibinfo{title}{Detecting natural disasters, damage, and incidents in the wild}.
\newblock In \emph{\bibinfo{booktitle}{The European Conference on Computer Vision (ECCV)}} (\bibinfo{year}{2020}).

\bibitem{zhu2020msnet}
\bibinfo{author}{Zhu, X.}, \bibinfo{author}{Liang, J.} \& \bibinfo{author}{Hauptmann, A.}
\newblock \bibinfo{journal}{\bibinfo{title}{Msnet: A multilevel instance segmentation network for natural disaster damage assessment in aerial videos}}.
\newblock {\emph{\JournalTitle{arXiv preprint arXiv:2006.16479}}}  (\bibinfo{year}{2020}).

\bibitem{fujita2017damage}
\bibinfo{author}{Fujita, A.} \emph{et~al.}
\newblock \bibinfo{title}{Damage detection from aerial images via convolutional neural networks}.
\newblock In \emph{\bibinfo{booktitle}{2017 Fifteenth IAPR International Conference on Machine Vision Applications (MVA)}}, \bibinfo{pages}{5--8} (\bibinfo{organization}{IEEE}, \bibinfo{year}{2017}).

\bibitem{doshi2018satellite}
\bibinfo{author}{Doshi, J.}, \bibinfo{author}{Basu, S.} \& \bibinfo{author}{Pang, G.}
\newblock \bibinfo{journal}{\bibinfo{title}{From satellite imagery to disaster insights}}.
\newblock {\emph{\JournalTitle{arXiv preprint arXiv:1812.07033}}}  (\bibinfo{year}{2018}).

\bibitem{cosmiqworksnvidia}
\bibinfo{author}{CosmiQWorks, D.}
\newblock \bibinfo{title}{Nvidia: Spacenet on amazon web services (aws) datasets: The spacenet catalog}.

\bibitem{demir2018deepglobe}
\bibinfo{author}{Demir, I.} \emph{et~al.}
\newblock \bibinfo{title}{Deepglobe 2018: A challenge to parse the earth through satellite images}.
\newblock In \emph{\bibinfo{booktitle}{2018 IEEE/CVF Conference on Computer Vision and Pattern Recognition Workshops (CVPRW)}}, \bibinfo{pages}{172--17209} (\bibinfo{organization}{IEEE}, \bibinfo{year}{2018}).

\bibitem{kyrkou2019deep}
\bibinfo{author}{Kyrkou, C.} \& \bibinfo{author}{Theocharides, T.}
\newblock \bibinfo{title}{Deep-learning-based aerial image classification for emergency response applications using unmanned aerial vehicles.}
\newblock In \emph{\bibinfo{booktitle}{CVPR Workshops}}, \bibinfo{pages}{517--525} (\bibinfo{year}{2019}).

\bibitem{rudner2019multi3net}
\bibinfo{author}{Rudner, T.~G.} \emph{et~al.}
\newblock \bibinfo{title}{Multi3net: segmenting flooded buildings via fusion of multiresolution, multisensor, and multitemporal satellite imagery}.
\newblock In \emph{\bibinfo{booktitle}{Proceedings of the AAAI Conference on Artificial Intelligence}}, vol.~\bibinfo{volume}{33}, \bibinfo{pages}{702--709} (\bibinfo{year}{2019}).

\bibitem{chowdhury2020comprehensive}
\bibinfo{author}{Chowdhury, T.}, \bibinfo{author}{Rahnemoonfar, M.}, \bibinfo{author}{Murphy, R.} \& \bibinfo{author}{Fernandes, O.}
\newblock \bibinfo{title}{Comprehensive semantic segmentation on high resolution uav imagery for natural disaster damage assessment}.
\newblock In \emph{\bibinfo{booktitle}{2020 IEEE International Conference on Big Data (Big Data)}}, \bibinfo{pages}{3904--3913} (\bibinfo{organization}{IEEE}, \bibinfo{year}{2020}).

\bibitem{christie2018functional}
\bibinfo{author}{Christie, G.}, \bibinfo{author}{Fendley, N.}, \bibinfo{author}{Wilson, J.} \& \bibinfo{author}{Mukherjee, R.}
\newblock \bibinfo{title}{Functional map of the world}.
\newblock In \emph{\bibinfo{booktitle}{Proceedings of the IEEE Conference on Computer Vision and Pattern Recognition}}, \bibinfo{pages}{6172--6180} (\bibinfo{year}{2018}).

\bibitem{michael}
\bibinfo{author}{Fernandes, O.} \emph{et~al.}
\newblock \bibinfo{title}{Quantitative data analysis: Small unmanned aerial systems at hurricane michael}.
\newblock In \emph{\bibinfo{booktitle}{2019 IEEE International Symposium on Safety, Security, and Rescue Robotics (SSRR)}}, \bibinfo{pages}{116--117}, \url{10.1109/SSRR.2019.8848935} (\bibinfo{year}{2019}).

\bibitem{V7Darwin}
\bibinfo{title}{V7 darwin}.
\newblock \bibinfo{howpublished}{\url{https://www.v7labs.com/darwin}}.
\newblock \bibinfo{note}{Accessed: 2020-08-25}.

\bibitem{fema2020fema}
\bibinfo{author}{FEMA}.
\newblock \bibinfo{title}{Fema preliminary damage assessment guide}.
\newblock \bibinfo{howpublished}{\url{https://www.fema.gov/sites/default/files/2020-07/fema_preliminary-disaster-assessment_guide.pdf}} (\bibinfo{year}{2020}).

\bibitem{rahnemoonfar:rescuenet2023}
\bibinfo{author}{Rahnemoonfar, M.}, \bibinfo{author}{Chowdhury, T.} \& \bibinfo{author}{Murphy, R.~R.}
\newblock \bibinfo{title}{Rescuenet: A high resolution uav semantic segmentation dataset for natural disaster damage assessment}.
\newblock \bibinfo{howpublished}{\emph{figshare} \url{https://doi.org/XXXXX}} (\bibinfo{year}{2023}).

\bibitem{zhao2017pyramid}
\bibinfo{author}{Zhao, H.}, \bibinfo{author}{Shi, J.}, \bibinfo{author}{Qi, X.}, \bibinfo{author}{Wang, X.} \& \bibinfo{author}{Jia, J.}
\newblock \bibinfo{title}{Pyramid scene parsing network}.
\newblock In \emph{\bibinfo{booktitle}{Proceedings of the IEEE conference on computer vision and pattern recognition}}, \bibinfo{pages}{2881--2890} (\bibinfo{year}{2017}).

\bibitem{chen2018encoder}
\bibinfo{author}{Chen, L.-C.}, \bibinfo{author}{Zhu, Y.}, \bibinfo{author}{Papandreou, G.}, \bibinfo{author}{Schroff, F.} \& \bibinfo{author}{Adam, H.}
\newblock \bibinfo{title}{Encoder-decoder with atrous separable convolution for semantic image segmentation}.
\newblock In \emph{\bibinfo{booktitle}{Proceedings of the European conference on computer vision (ECCV)}}, \bibinfo{pages}{801--818} (\bibinfo{year}{2018}).

\bibitem{strudel2021segmenter}
\bibinfo{author}{Strudel, R.}, \bibinfo{author}{Garcia, R.}, \bibinfo{author}{Laptev, I.} \& \bibinfo{author}{Schmid, C.}
\newblock \bibinfo{title}{Segmenter: Transformer for semantic segmentation}.
\newblock In \emph{\bibinfo{booktitle}{Proceedings of the IEEE/CVF international conference on computer vision}}, \bibinfo{pages}{7262--7272} (\bibinfo{year}{2021}).

\bibitem{oktay2018attention}
\bibinfo{author}{Oktay, O.} \emph{et~al.}
\newblock \bibinfo{journal}{\bibinfo{title}{Attention u-net: Learning where to look for the pancreas}}.
\newblock {\emph{\JournalTitle{arXiv preprint arXiv:1804.03999}}}  (\bibinfo{year}{2018}).

\bibitem{chen2017rethinking}
\bibinfo{author}{Chen, L.-C.}, \bibinfo{author}{Papandreou, G.}, \bibinfo{author}{Schroff, F.} \& \bibinfo{author}{Adam, H.}
\newblock \bibinfo{journal}{\bibinfo{title}{Rethinking atrous convolution for semantic image segmentation}}.
\newblock {\emph{\JournalTitle{arXiv preprint arXiv:1706.05587}}}  (\bibinfo{year}{2017}).

\bibitem{ronneberger2015u}
\bibinfo{author}{Ronneberger, O.}, \bibinfo{author}{Fischer, P.} \& \bibinfo{author}{Brox, T.}
\newblock \bibinfo{title}{U-net: Convolutional networks for biomedical image segmentation}.
\newblock In \emph{\bibinfo{booktitle}{International Conference on Medical image computing and computer-assisted intervention}}, \bibinfo{pages}{234--241} (\bibinfo{organization}{Springer}, \bibinfo{year}{2015}).

\end{thebibliography}

\end{document}